\documentclass[twocolumn]{article}
\usepackage{gensymb}
\usepackage{siunitx,comment}
\usepackage{graphicx}%
\usepackage{hyperref}
\usepackage{amsmath,amssymb,amsfonts}%
\usepackage{amsthm}%
\usepackage{mathrsfs}%
\usepackage[title]{appendix}%
\usepackage{xcolor}%
\usepackage{textcomp}%
\usepackage{manyfoot}%
\usepackage{booktabs}%
\usepackage{algorithm}%
\usepackage{algorithmicx}%
\usepackage{algpseudocode}%
\usepackage{listings}
\theoremstyle{thmstyleone}%

\usepackage[utf8]{inputenc}
\usepackage[T1]{fontenc}
\usepackage{xcolor}
\usepackage{color,soul}
\theoremstyle{thmstyletwo}%

\theoremstyle{thmstylethree}%

\title{Deep Learning  Based Illegal Bowling Action Detection}
\author{Debopom Sutradhar$^1$ \thanks{Correspondence: dsutradhar201046@bscse.uiu.ac.bd}, Niful Islam$^1$, Sudipto Mondal$^1$, Tasmima Hossain Jamim$^1$, \\Jubaer Muhammad Shufol$^1$ and Swakkhar Shatabda$^2$ \\
$^1$Department of Computer Science and Engineering, United International University \\%Dhaka, Bangladesh\\
$^2$Department of Computer Science and Engineering, BRAC University \\%Dhaka, Bangladesh
}
\date{}
\begin{document}
\maketitle

\abstract{Cricket, often referred to as the "gentleman's game," adheres to a strict rule set for both batsmen and bowlers, where each delivery can significantly impact the match outcome. Detecting illegal bowling actions is crucial for maintaining fair play, yet it remains challenging for umpires to monitor in real time. Existing sensor-based solutions have limitations in live match scenarios, making real-time assessment difficult. This paper proposes a computer vision-based deep learning solution to detect illegal bowling actions in live cricket matches. To develop and evaluate our approach, we compiled a dataset of 62 videos featuring 11 male bowlers, capturing both legal and illegal bowling actions from multiple angles—front, back, and side. However, the dataset predominantly comprises right-handed bowlers with conventional actions.} The proposed system identifies two key frames, the shoulder frame and the release frame from video footage of a bowler’s delivery and analyzes the change in the bowling arm's angle between these frames. If the angle difference exceeds a predefined threshold (e.g., 15 degrees), the delivery is flagged as potentially illegal. We evaluated the system on a custom dataset and achieved a high true positive rate, suggesting the system’s potential effectiveness in real-time match settings. However, further research is required to validate the system across diverse environmental conditions and larger datasets to ensure generalizability and robustness in various live match scenarios. To the best of our knowledge, this is the first AI-based computer vision method for detecting illegal bowling actions in cricket.

%\keywords{Bowling, Cricket, Deep Learning, Pose Estimation, Object Detection}


\section{Introduction}
Cricket, one of the most popular sports worldwide, has a rich history, complex techniques, and an exhilarating rivalry between bat and ball. Bowling, a pivotal aspect of the game, can significantly influence a match, where a single delivery has the potential to change the course of play. Illegal bowling actions, where a bowler's arm is bent beyond an allowed threshold during delivery, have long been a contentious issue in cricket \cite{muddassar2020bowling}. This rule is in place to ensure fairness and uphold the spirit of the game. In the past, some bowlers have sought to gain an unfair advantage by altering their arm movements to increase pace or spin, making the detection of illegal actions crucial for maintaining integrity. However, real-time detection of these actions remains challenging, as current methods, which rely on wearable sensors, involve a lengthy 21-day analysis process conducted post-match \cite{spratford2018illegal}.

Various computer vision methods have been proposed in cricket for applications such as detecting no-balls by tracking the bowler's front foot at the crease \cite{chowdhury2016application}, assisting umpires with Decision Review Systems (DRS) \cite{shirkedrs}, and even detecting the delivery length through deep learning \cite{abbas2022deep}. Other advancements in computer vision include pose estimation techniques, like generating heatmaps for joint localization \cite{artacho2020unipose} and top-down pose estimation using a cascade transformer to predict key pose points \cite{li2021pose}. Despite these developments, no computer vision approaches have been specifically applied for the real-time detection of illegal bowling actions during live matches. Given that current laboratory-based detection methods are resource-intensive, time-consuming, and impractical for in-game use, there is a pressing need for a real-time, computer vision-based system capable of identifying illegal bowling actions efficiently.

Computer vision, a branch of artificial intelligence, enables systems to interpret visual data, transforming images and videos into meaningful information for various applications. In healthcare, for instance, it plays an essential role in medical image analysis, aiding in early disease detection. Additionally, computer vision is used in tasks such as object recognition, image classification, pose estimation, and augmented reality. In the context of cricket, computer vision has facilitated applications like no-ball detection, DRS support, and delivery length estimation \cite{chowdhury2016application, shirkedrs, abbas2022deep}. 

Pose estimation is another relevant area in computer vision, where systems detect key points on a person or object to infer poses from images or videos. Applications of pose estimation span human activity recognition, object tracking, augmented reality, and robotics, among others \cite{zheng2023deep, meinhardt2022trackformer, buchner2023media}. In this paper, we leverage pose estimation and other computer vision techniques to propose a novel system for detecting illegal bowling actions in real time, with potential applications in live broadcast analytics, automated umpiring, and post-match performance evaluation. Our classifier identifies two specific frames within a bowling video: the shoulder frame and the release frame. Using MediaPipe to detect the shoulder frame and RT-DETR for release frame extraction, we then calculate the difference in the bowling arm's angle between these frames. If the angle difference exceeds \SI{15}{\degree}, the delivery is flagged as illegal. 

This study introduces a labeled dataset of 62 videos across two classes, legal and illegal bowling actions to evaluate the system's performance. The key contributions of this paper include:
\begin{itemize}
    \item To the best of our knowledge, we are the first to develop a deep learning-based approach for automatically detecting illegal bowling actions in live cricket matches. \item We have curated the first dataset of labeled legal and illegal bowling actions, comprising 62 videos of 11 bowlers, which we use to evaluate our solution. 
    \item Experimental analysis shows that our solution achieved an accuracy of 91.96\% and a true positive rate of 100\% on a controlled dataset, suggesting high potential for effective performance.
    \item We have made our dataset and source code publicly available to support future research \cite{github_repo}.
\end{itemize}

% While our results are promising, further research is necessary to validate the system across different environments and with larger datasets to enhance its robustness and generalizability in diverse live match conditions.

The rest of the paper is organized as follows. Section \ref{sec:preli} holds a detailed overview of the preliminary details. Section \ref{sec:rw} holds a brief overview of the research works along with a demonstration of the research gap. Section \ref{sec:method} holds the proposed model, followed by the results in Section \ref{sec:result}. The article terminates in Section \ref{sec:conclu} discussing the limitations of the work and future scope of improvement.

\section{Preliminaries}
\label{sec:preli}
This section covers the fundamental concepts of ICC rules of illegal bowling action, pose estimation, and object detection.

\subsection{ICC Rules for Bowling Action}

\begin{figure}
    \centering
    \includegraphics[height=8cm, width=.7\linewidth]{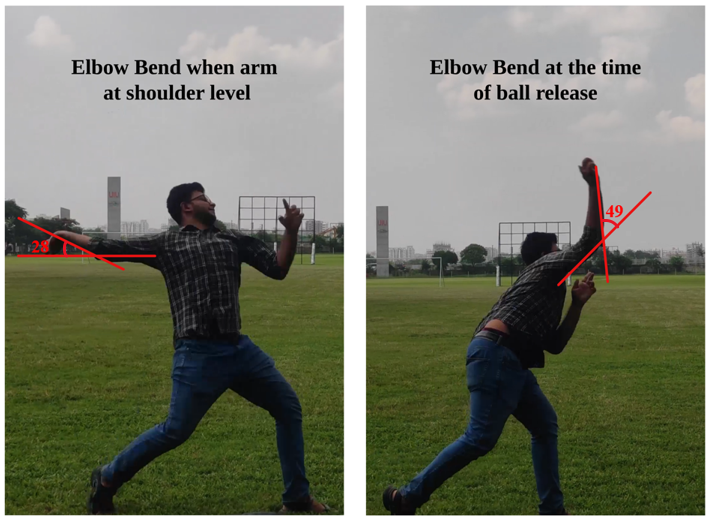}
    \caption{ICC rule on illegal bowling action}
    \label{fig:15dr}
\end{figure}
% add diagram + description + cite -> shufol 
% grammar check
Chucking is the formal name of throwing in cricket. According to the rule of International Cricket Council (ICC), chucking is illegal since it gives an extra advantage to the bowler.
A bowl is considered chucking if the angle difference between the elbow bend at the shoulder level and the elbow bend at the time of release is  more than \SI{15}{\degree} the time of releasing the bowl. Figure \ref{fig:15dr} visually represents the elbow bend of shoulder level and elbow bend of release point.
If a bowler is suspected of chucking by the on field umpire. The umpire can call no ball for the first occurrence. If chucking repeated again by the same bowler then umpire might warn the bowling team captain and bowler. For third occurrence the umpire might suspend the bowler . If the batsman find suspicious bowling, batsman can appeal for legality check to respective cricket board. Later the umpires inform ICC and respective cricket board about the incident and necessary action follow against the individual and must undergo through assesstment. For further assessment, the bowler will be tested in a laboratory mentioned by ICC. If the bowler if found guilty for chucking, he is banned from international cricket until he improves his bowling action. In a cricket match, a bowler can bowl a delivery without bending less than \SI{15}{\degree}. If a bowler is suspected of bowling an illegal delivery which is called chucking. Then he/she has to go through sensor tests provided by ICC.

The ICC has strict protocols for reporting suspected illegal deliveries in cricket by umpires via post-match reports. This leads to a bio-mechanical analysis of the bowler's action in an approved lab, where the bowler is required to bowl a series of deliveries monitored with motion capture, radar, and high-speed video. The analysis is conducted by a team of 5 people and takes 21 days for the data to be interpreted and reported by the biomechanist. The sensors used in the analysis consist of a combination of a $\pm$16g 3D accelerometer and a $\pm$2000 deg/s 3D gyroscope connected by a flexible PCB carrying power and an I2C data bus \cite{wixted2011detection}. 

\subsection{Pose Estimation}
%  no itemize + diagram -> dipto
A computer vision task called pose estimation infers a person or object's pose from an image or video. This technique first tracks a human from an image or video. Then different types of landmarks are put on different joints of the human body. After connecting those landmarks, a simplified human skeleton is made. By that skeleton, the pose of that human can be determined \cite{josyula2021review}. %Pose estimation can alternatively be defined as the challenge of figuring out the location and orientation of a camera in relation to a specific person or item. 
MediaPipe \cite{lugaresi2019mediapipe} is a framework widely used for building real-time multimedia processing pipelines. There are several tasks to be performed in our research. First task is to detect the two frames on which the bowler's bowling arm is in shoulder label and the frame on which the bowler is releasing the bowl. The second task to be performed with MediaPipe is to detect the angle between arms. If the difference of the angle in the above two frames is greater than \SI{15}{\degree}, it is considered as an illegal bowling action.

\subsection{Object Detection}
Object detection  is typically used for locating instances of objects in images or videos \cite{zou2023object}. You Only Look Once (YOLO) \cite{redmon2016you} is a state-of-the-art object detection algorithm. It is a real-time object detection system that uses a single neural network to predict bounding boxes and class probabilities for objects in an image. 
\section{Related Work}
\label{sec:rw}

%Several significant topics in the identification of illegal bowling action in cricket technology are presented in the associated work section. Initially, using video data analysis, it investigates automatic no-ball detection. Furthermore, it explores how computer vision is used in cricket umpiring, with a focus on advanced decision-making systems. When pitch analysis and object identification are combined, an additional topic shows how they work together to improve cricket scenarios. The section concludes with a discussion of comparative techniques that use the CNN with SVM and YOLO algorithms to improve cricket video analysis. When taken as a whole, these features highlight the variety of technical advancements used to improve several areas of the cricketing experience.
This section provides a brief summary of the literature on video based methods, computer vision based umpiring, automated pitch analysis and comparative methods on the game of cricket. 
\subsection{Video based methods}
Chowdhury et al. \cite{chowdhury2016application} proposed a method to detect bowler overstepping using video analysis. A base frame is selected before the game, and the popping crease is identified using the Hough technique. The crease divides the scene into two regions: a no-ball region (left of the crease) and a lawful region. Frame subtraction compares subsequent frames to the base frame, detecting pixel changes. Mean changes in both regions are calculated, with a predefined threshold determining no-ball conditions. If the mean for the no-ball region exceeds its threshold and the lawful region's mean is below its threshold, a no-ball is detected; otherwise, the delivery is legal.

\subsection{Vision based umpiring}
Rashid et al. \cite{harun2018crick} detected waist-high no-balls using InceptionV3, training the final layer on a dataset of 1000 images with K-fold cross-validation for accuracy. Their system aids umpires in making fair decisions for critical moments like LBW, no-balls, and run-outs using computer vision. The 2D coordinates of the tracked ball, obtained via the Minimum Enclosing Circle algorithm, are mapped to 3D world coordinates based on pitch dimensions. Linear and quadratic regression help visualize the ball and pitch, enhancing decision-making in training and international matches \cite{shirkedrs}.

Karmaker et al. \cite{karmaker2015cricket} classified cricket shots into square cuts, hooks, flicks, and off drives. They used the Laplacian of Gaussian (LoG) for edge detection, determined motion vectors using intensity values, and classified shots based on pre-defined angles. Abbas et al. \cite{abbas2022deep} focused on delivery length determination. They extracted delivery clips from videos, tracked umpires and pitch using YOLO, identified the ball with RetinaNet, and calculated bounce location and delivery segmentation for analysis.

\subsection{Automated Pitch Analysis}
To detect the batsman and bowler using the ResNet50 model, first identify their positions in the frame and calculate the batsman's height and the pitch height from the batsman's bottom to the bowler's bottom. In the second frame, detect the batsman again, compute the pitch height using a formula, and map the ball's bounce position to categorize the delivery type. For third umpire decisions and automated scoring, CNN and Inception V3 are used for classification, with deep CNN enhancing accuracy. Image subtraction algorithms analyze the bowling crease to determine no-balls or legal deliveries, while Hawk Eye evaluates bowling turning. Datasets, including umpire signals, are resized to 227x227x3 and augmented with transformations like rotation and flipping. PCA or ZCA whitening is applied for dimensionality reduction, with ZCA preserving core image resemblance. This integrated approach leverages advanced algorithms and datasets to improve decision-making and automate cricket scoring \cite{kowsher2019detecting}.

\subsection{Comparative methods}
Shukla et al. \cite{shukla2018automatic} proposed a model for extracting highlights from cricket videos using a CNN+SVM approach. They trained a binary linear SVM on features extracted from AlexNet's fc7 layer to classify keyframes as starting frames. Utilizing the IPL dataset with 15 hours of video, they identified events like boundaries, wickets, and celebrations by analyzing the scoreboard, and generated highlights accordingly. Guntuboina et al. \cite{guntuboina2021deep} introduced a method to extract key match points by detecting scoreboards using the YOLO algorithm, which treats the problem as a regression for faster processing. They split the data into 80\% for training and 20\% for testing, extracting 30 images per second and cropping scoreboard sections for model training. This approach efficiently extracts critical match moments from videos, focusing on four shot types: pitch view, close-up view, audience view, and medium view. Shots are classified based on features like color, texture, motion, and edges using a Hidden Markov Model (HMM), with highlights compiled from keyframes. However, the method does not address chucking detection or pose estimation in bowling actions.

Table \ref{tab:gap-analysis} presents a comparative analysis of different methods, making it clear that existing research works do not focus on computer vision-based approaches for detecting illegal bowling actions. While some methods do attempt to identify illegal bowling actions, they often require additional hardware equipment, such as wearable sensors. However, these solutions are not feasible for real-match scenarios. To address this gap, we propose a computer vision-based solution capable of effectively distinguishing illegal bowling actions in real time without requiring any extra hardware. This approach not only enhances rule enforcement and protects the integrity of cricket but also advances research in computer vision for sports analytics.

\begin{table*}[h]
    \caption{Comparative analysis of different solutions}
    \label{tab:gap-analysis}
    \centering
    \begin{scriptsize}
    \begin{tabular}{|p{3cm}|p{1.2cm}|p{1.2cm}|p{1.2cm}|p{1.2cm}|p{1.2cm}|p{1.2cm}|p{1.2cm}|}
    \hline
        \textbf{Approach} & \textbf{Legal Bowling Detection} & \textbf{No ball 
        Detection} & \textbf{Out Detection} & \textbf{Sensor used} & \textbf{Video Analysis} & \textbf{Computer Vision Techniques Applied} & \textbf{Pose Estimation} \\ \hline
        A Wearable Wireless Sensor based \cite{ahmed2015wearable} & Yes & No & No & Yes & No & No & No \\ \hline
        Inertial sensor orientation \cite{wixted2011inertial} & Yes & No & No & No & No & No & No \\ \hline
        Application of Computer Vision \cite{chowdhury2016application} & No & Yes & No & No & Yes & Yes & No \\ \hline
        Crick-net: A Convolutional Neural Network based Classification \cite{harun2018crick} & No & Yes & No & No & Yes & Yes & No \\ \hline
        DRS Assisting Umpire Using Computer Vision \cite{shirkedrs} & No & Yes & Yes & No & Yes & Yes & No \\ \hline
        Multi-person 3D Pose Estimation \cite{bridgeman2019multi} & No & No & No & No & Yes & Yes & Yes \\ \hline
        An evaluation of biomechanical measures \cite{ferdinands2007evaluation} & Yes & No & No & No & Yes & No & No \\ \hline
        Automatic Cricket Highlight generation \cite{shukla2018automatic} & No & No & No & No & Yes & Yes & No \\ \hline
        AI Coach: Deep Human Pose Estimation \cite{wang2019ai} & No & No & No & No & Yes & Yes & Yes \\ \hline
        Multi-person 3D Pose Estimation \cite{bridgeman2019multi} & No & No & No & No & Yes & Yes & Yes \\ \hline
    \end{tabular}
        \end{scriptsize}
    
\end{table*}

\section{Proposed System}
\begin{figure*}[h]
    \centering
    \includegraphics[width=1.0\linewidth]{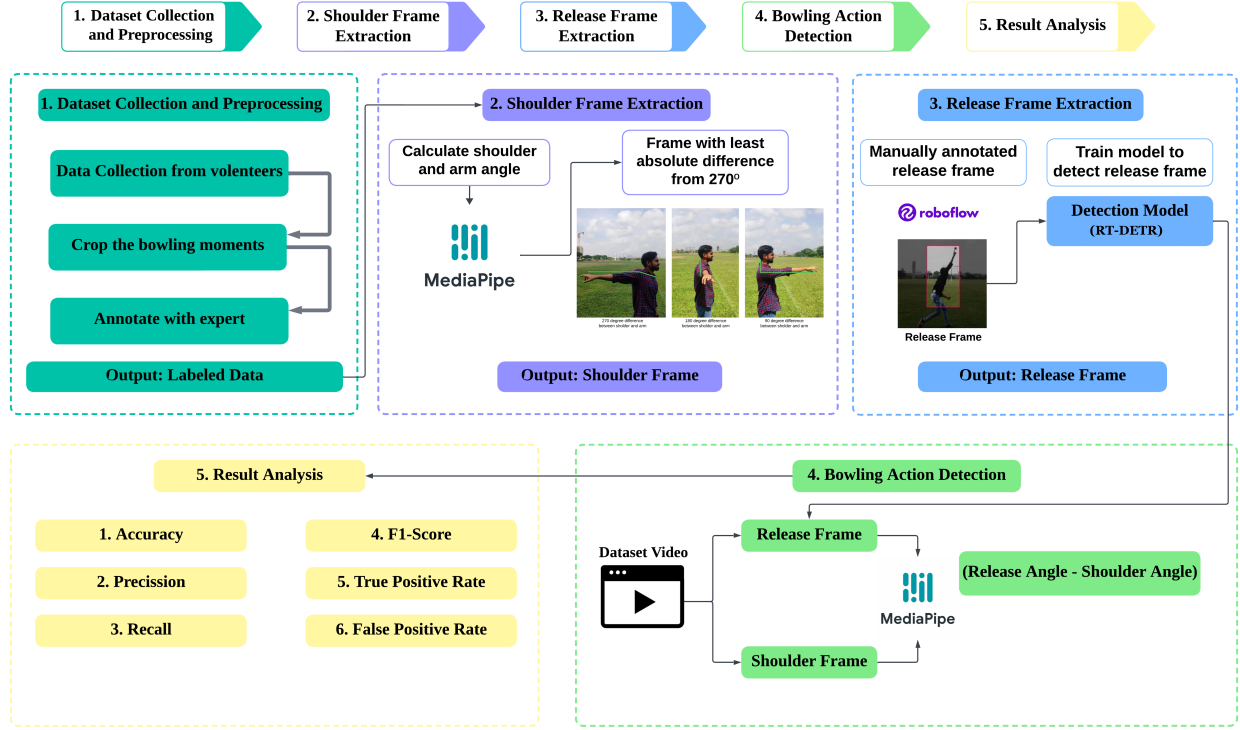}
    \caption{Workflow diagram of the whole process}
    \label{fig:workflow}
\end{figure*}
\label{sec:method}
This section describes the proposed method in detail. The proposed system has four major steps. These are dataset collection and preprocessing, shoulder frame extraction, release frame extraction, and decision-making. Figure \ref{fig:workflow} depicts the workflow in detail. Firstly we collect and preprocess the data thoroughly. In the second and third stages, we extract the shoulder frame and the release frame from the bowling video respectively. In the fourth stage, we analyze the extracted frames for final decision-making.

\subsection{Dataset}
\begin{figure}[h] 
    \centering
    \includegraphics[width=0.8\linewidth]{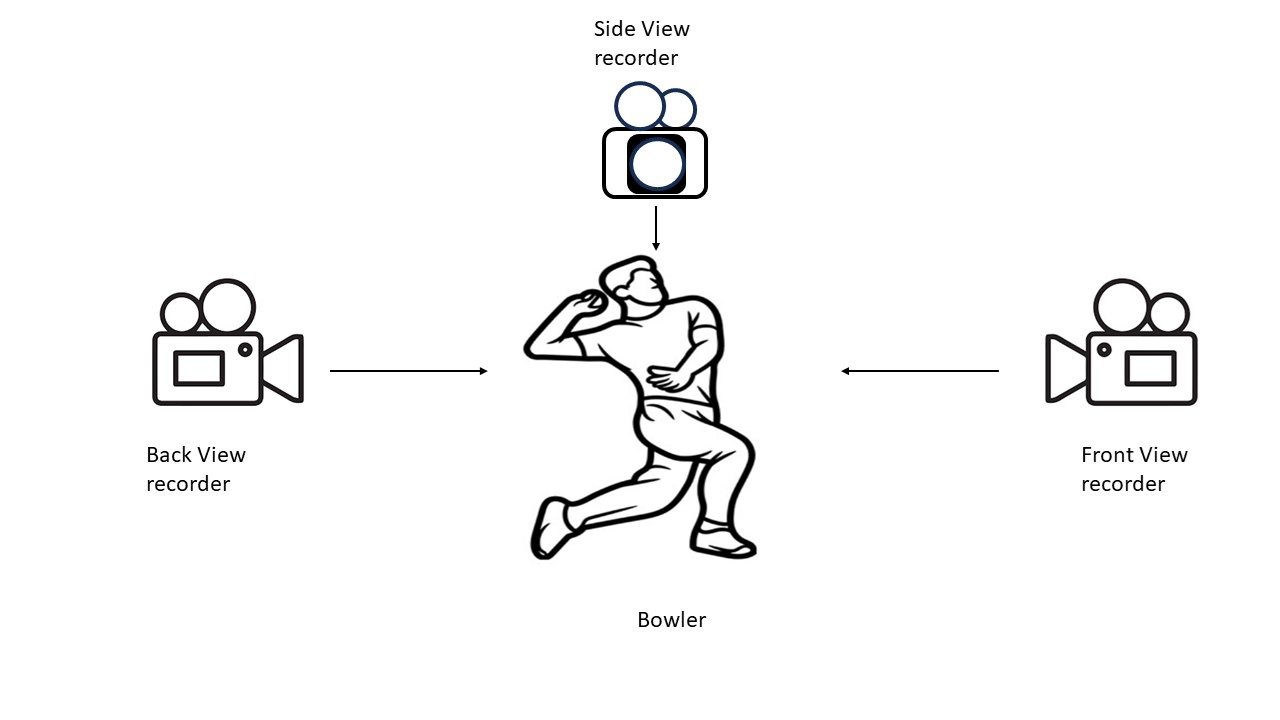}
    \caption{Camera angle of bowling data collection}
    \label{fig:camera}
\end{figure}

Bowling video footage documenting both legal and illegal bowling incidents has been methodically assembled from volunteers. The video is captured from three different angles: front, back, and side to ensure a large and complete dataset available for analysis. The overview of the data-collecting procedure is shown in Figure \ref{fig:camera}. The selection of 11 bowlers was based on the availability of participants who could provide diverse bowling actions while maintaining a controlled dataset for analysis. This sample size, though limited, allows for an initial assessment of the system’s effectiveness in detecting illegal actions. The dataset includes multiple deliveries per bowler, ensuring a range of variations within individual bowling styles. The legality of the bowling action is first established by looking at the side view. The front and rear views are utilized when the expert is having trouble assessing the legality. These other perspectives provide insightful information on the bowlers' motions, allowing for a more thorough assessment of their methods. Through the integration of many perspectives, the data collection procedure guarantees a thorough examination of both legal and illegal bowling actions. Due to the methodical approach, experts and researchers are able to analyze the tactics that players use in detail, which leads to a better comprehension of the difficulties associated with judging whether bowling acts are lawful.

We have collected a total of 62 videos of 11 bowlers of which 36 are labeled by the experts. A subset of 12 labelled data instances is kept aside for the pipeline's testing phase. The test set is rigorously selected in accordance with stringent testing processes to ensure that each bowler, having provided multiple deliveries, is exclusively allotted to the test set, hence precluding their inclusion in the training set. Since training YOLOV8 does not require labelled instances, unlabeled videos are used to train the model. Table \ref{tab:data_meta} holds the data collection description, which includes key statistics about the dataset used in our analysis. During the preprocessing stage, each video is cropped to keep only the parts corresponding to bowling moments.

\begin{table}[h]
    \centering
    \caption{Data collection description}
    \begin{tabular}{|l|l|}
    \hline
        \textbf{Description} & \textbf{Value} \\ \hline
        Total number of videos & 62 \\ \hline
        Total number of bowlers & 11 \\ \hline
        Videos for testing & 12 \\ \hline
        Number of unique bowlers in test set & 4 \\ \hline
        Number of videos for training & 50 \\ \hline
    \end{tabular}
    \label{tab:data_meta}
\end{table}

\subsection{Shoulder Frame Extraction}
\begin{figure}
    \centering
    \includegraphics[width=\linewidth]{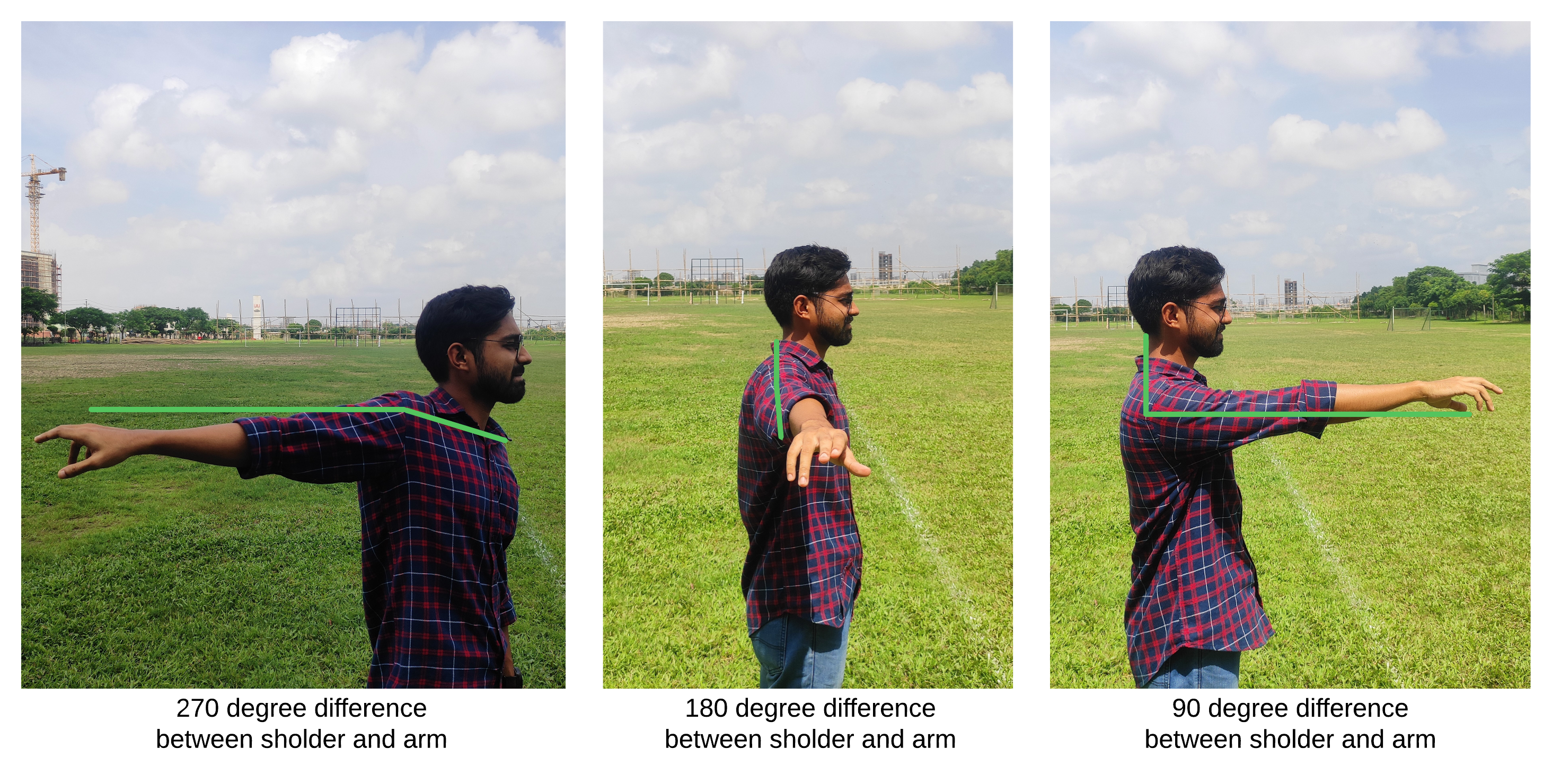}
    \caption{Angle of different gestures with the shoulder}
    \label{fig:sholder_frame}
\end{figure}

As presented in Figure \ref{fig:sholder_frame}, in the ideal case, the hand makes a \SI{270}{\degree} angle with the shoulders while bowling. Therefore, the frame having the least absolute difference from the expected angle is marked as the shoulder frame. Let $F_i$ represent the frame index where the hand's angle with the shoulder is closest to \SI{270}{\degree}. To detect this frame, as shown in Equation \ref{eq:sholder}, we iterate every frame, $A_i$, and select the one with the least absolute difference from \SI{270}{\degree}. We have leveraged Mediapipe to measure the angle. 

\begin{footnotesize}
\begin{equation}
\label{eq:sholder}
F_i = \arg\min_{i} |A_i - 270|
\end{equation}
\end{footnotesize}

% equation
\subsection{Release Frame Extraction}
\begin{figure} [h]
    \centering
    \includegraphics[width=\linewidth]{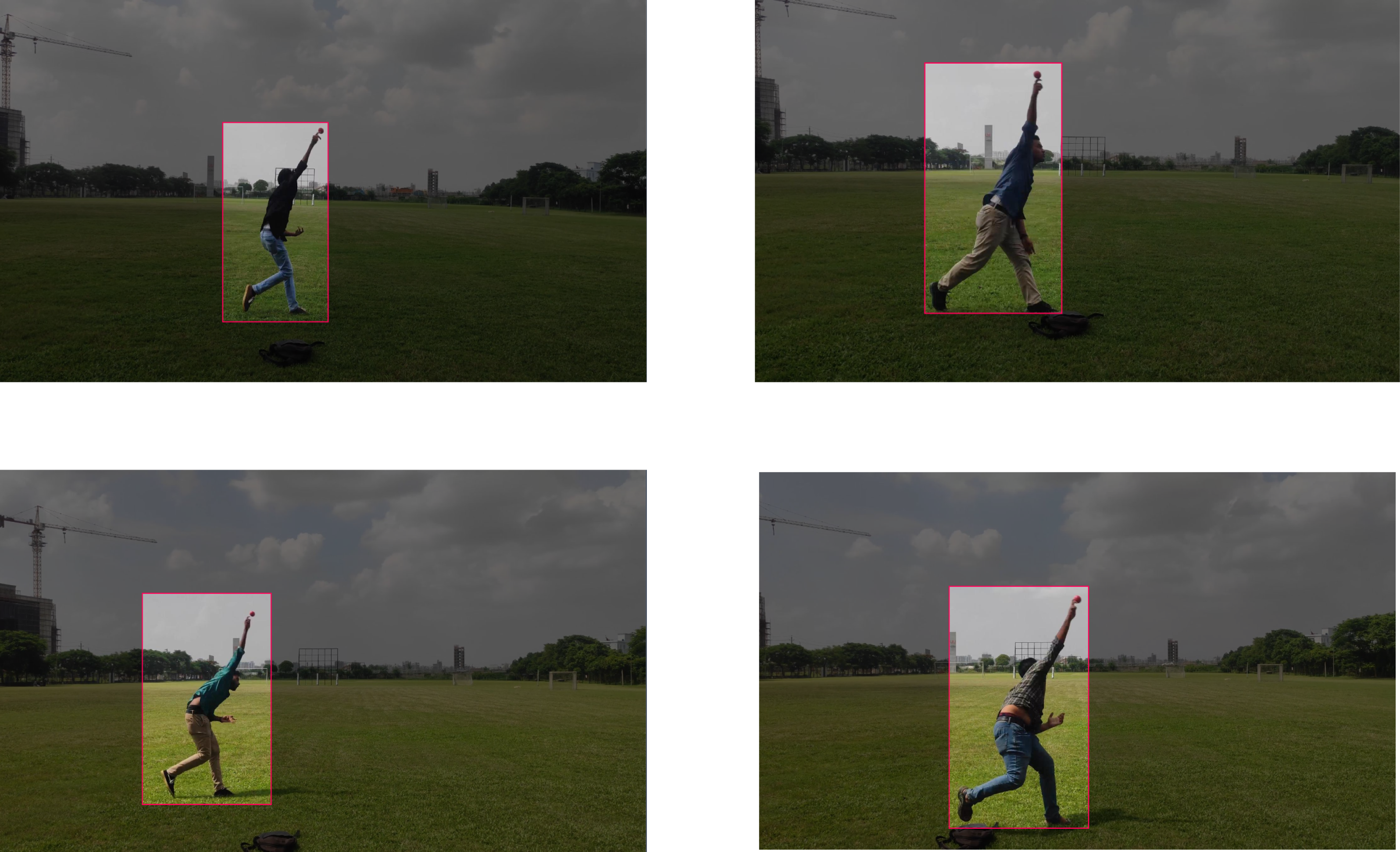}
    \caption{Annotated data by roboflow}
    \label{fig:robo}
\end{figure}

In this section, a particular frame where the bowler releases the ball during the bowling action is extracted from the video. We have used Roboflow\footnote[1]{https://roboflow.com/} to manually select the release frame that is used for training and validation. Figure \ref{fig:robo} shows a visual representation of the annotated data with Roboflow. A YOLOV8 model is trained with the annotated data 
and it was chosen over YOLOv7 due to its improved backbone for feature extraction, enhanced neck architecture for better multi-scale feature fusion, anchor-free detection for streamlined object localization, and optimized loss functions and post-processing. The checkpoints system ensures that the best-performing model on the validation data is kept. The frame with the highest probability of being the release frame is returned as output from the stage.

\subsection{Bowling Action Detection}
\begin{figure} [h]
    \centering
    \includegraphics[width=\linewidth]{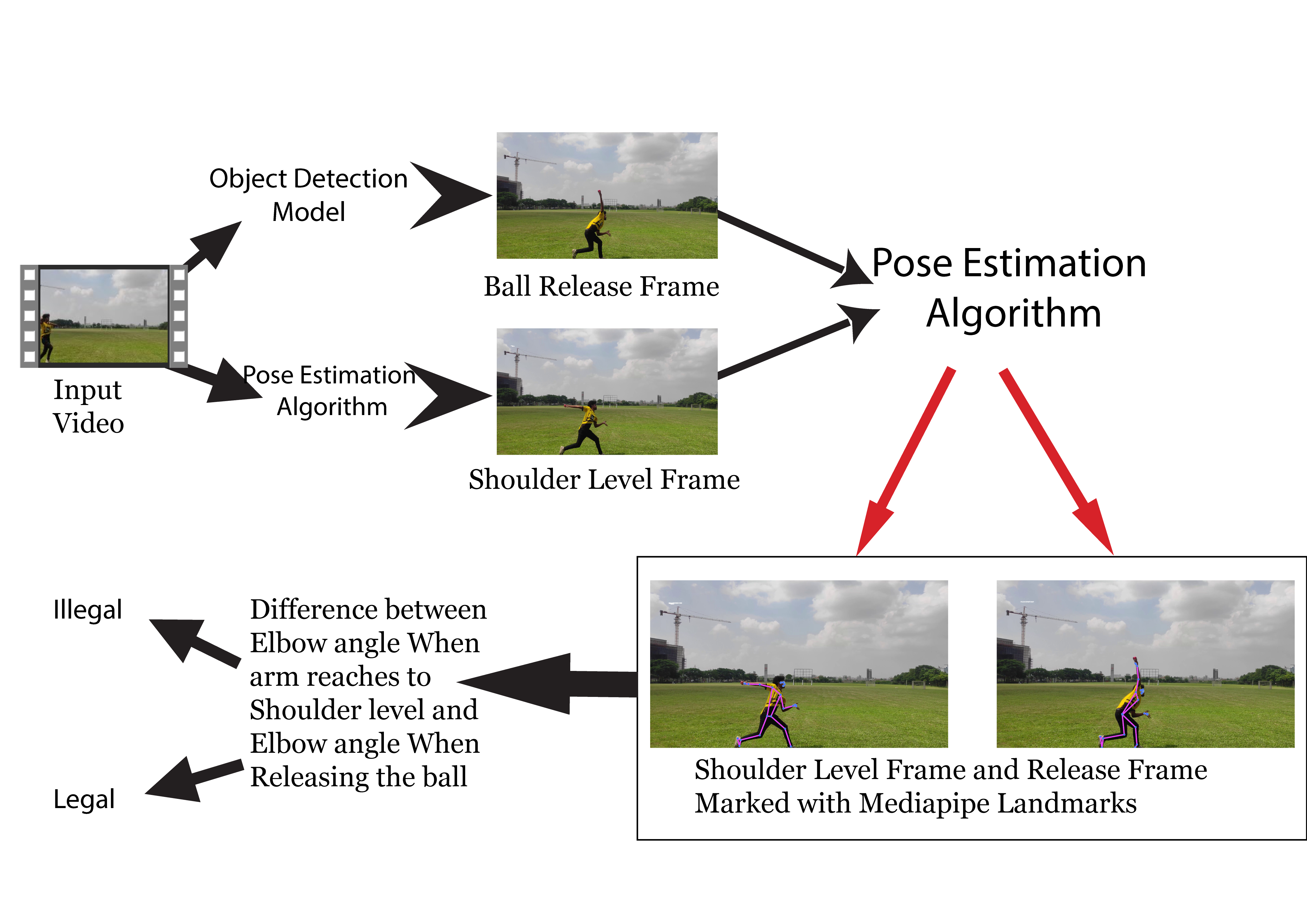}
    \caption{Workflow of the decision making}
    \label{fig:decision}
\end{figure}

For the final decision-making of bowling action, we have leveraged the previous two steps to extract the corresponding shoulder and release frames extraction. For the final decision making, only the two frames from the video are considered. This stage incorporates two steps. The first stage uses Mediapipe to quantify the elbow's flexion on the two frames. In the next stage, the absolute difference between the measured flexions is calculated. If the difference is above \SI{15}{\degree}, the bowling action is marked illegal. Otherwise, it is considered legal. \sethlcolor{yellow} According to ICC regulations, a bowling action is deemed illegal if the elbow extension exceeds \SI{15}{\degree} \cite{wixted2011detection}. Figure \ref{fig:decision} shows the whole workflow of the decision-making. Firstly, the shoulder and the release frames are detected employing two algorithms discussed before. Subsequently, the angle of the bowling arm of the two frames is measured. If the difference is above the threshold, it is marked illegal.

\section{Experimental Results}
\label{sec:result}

This section presents the detailed methodology along with the experimental results. 

\subsection{Evaluation metrics}
We have tested our proposed solution with various evaluation metrics. These are Accuracy, Precision, Recall, F1-score, True Positive Rate, False Positive Rate and finally MCC. In each case, let the True Positive value be denoted by TP. In the same convention, the True Negative value is denoted by TN, False Positive by FP and False Negative by FN.

A classifier's accuracy on a given test set is the percentage of the correctly classified tuples amongst all tuples (TP, TN, FP, FN). The correctly classified tuples are the True Positive (TP) and the True Negative (TN) tuples.

\begin{footnotesize}
\begin{equation}
\text { Accuracy }=\frac{\mathrm{TP}+  \mathrm{TN}}{\mathrm{TP}+\mathrm{PP}+\mathrm{TN}+\mathrm{FN}}
\end{equation}
\end{footnotesize}

Precision gives the percentage of the predicted Positive instances were actually True Positives (TP) amongst all the positively labeled ones (TP and FP), that is, it measures the exactness of data set.

\begin{footnotesize}
\begin{equation}
\text { Precision }=\frac{\mathrm{TP}}{\mathrm{TP}+\mathrm{FP}}
\end{equation} 
\end{footnotesize}

Recall gives the True Positive rate, that is, the percentage of the positive instances that were predicted as such by the classifier.

\begin{footnotesize}
\begin{equation}
\text { Recall }=\frac{\mathrm{TP}}{\mathrm{TP}+\mathrm{FN}}
\end{equation}
\end{footnotesize}

 F1-score gives a single measure by combining Precision and Recall. The value of F1-score ranges from 0 to 1 where a higher value indicates a better performance.

\begin{footnotesize}
\begin{equation}
 \text { F1-score }=\frac{2 \times \text { Precision } \times \text { Recall }}{\text { Precision }+ \text { Recall }}
\end{equation}
\end{footnotesize}

True Positive Rate or Sensitivity gives the measurement of the proportion of Positive Tuples (TP and FP) that were labeled as True Positive (TP).

\begin{footnotesize}
\begin{equation}
\text{True Positive Rate} = \frac{TP}{TP + FN}
\end{equation}
\end{footnotesize}

False Positive Rate gives the measurement of the proportion of actual Negative tuples that were incorrectly labeled by the classifier.

\begin{footnotesize}
\begin{equation}
\text{False Positive Rate} = \frac{FP}{FP + TN}
\end{equation}
\end{footnotesize}

MCC or Matthews Correlation Coefficient combines all four outcomes, that is, the true positives, true negatives, false positives, and false negatives to provide an assessment of the classifier's performance. The values of MCC ranges from -1 to 1 where -1 indicates perfect inverse prediction and 1 indicates perfect prediction and 0 indicates random prediction.

\begin{footnotesize}
\begin{equation}
\mathrm{MCC}=\frac{\mathrm{TP} \times \mathrm{TN}-\mathrm{FP} \times \mathrm{FN}}{\sqrt{(\mathrm{TP}+\mathrm{FP})(\mathrm{TP}+\mathrm{FN})(\mathrm{TN}+\mathrm{FP})(\mathrm{TN}+\mathrm{FN})}}
\end{equation}
\end{footnotesize}

Each of the above-mentioned metrics provides a different perspective on the model's performance.

\subsection{Release Frame Extraction Result}
Table~\ref{tab:release_frame_models} presents the performance comparison of different object detection models using Accuracy and mean Average Precision (mAP). Among all evaluated models, RT-DETR achieved the highest performance, obtaining an accuracy of 94.6\% and an mAP of 95.0\%. These results demonstrate its superior ability to accurately localize the release frame compared with both one-stage and two-stage object detection approaches.

The YOLO family of detectors also produced competitive results, with performance improving consistently across successive versions. YOLOv5 achieved an accuracy of 85.9\%, which increased to 88.2\%, 89.4\%, 90.1\%, and 91.0\% for YOLOv7, YOLOv8, YOLOv10, and YOLOv11, respectively. This progression reflects the architectural improvements incorporated into newer YOLO variants for object localization and feature representation.

Traditional two-stage and single-stage detectors exhibited comparatively lower performance. Faster R-CNN, RetinaNet, and Mask R-CNN achieved accuracies ranging from 83.1\% to 85.0\%, while SSD produced the lowest accuracy of 79.8\%. Although these models were capable of detecting the release frame, they were less robust than the transformer-based RT-DETR in handling the variations present in the bowling videos.

Based on these experimental results, RT-DETR was selected as the release frame detector for the proposed bowling action detection pipeline. Its superior localization accuracy provides more reliable release frame extraction, thereby improving the precision of elbow angle estimation and contributing to the overall performance of the illegal bowling action detection system.

\begin{table}[h!]
\centering
\scriptsize
\caption{Release Frame Detection Model Comparison}
\label{tab:release_frame_models}
\begin{tabular}{lcc}
\hline
\textbf{Model} & \textbf{Accuracy (\%)} & \textbf{mAP (\%)} \\
\hline
YOLOv5 & 85.9 & 86.5 \\
YOLOv7 & 88.2 & 89.0 \\
YOLOv8 & 89.4 & 90.6 \\
YOLOv10 & 90.1 & 91.2 \\
YOLOv11 & 91.0 & 92.0 \\
Faster R-CNN & 83.1 & 84.0 \\
SSD & 79.8 & 81.0 \\
RetinaNet & 84.4 & 85.7 \\ 
Mask R-CNN & 85.0 & 86.3 \\
\textbf{RT-DETR} & \textbf{94.6} & \textbf{95.0} \\
\hline
\end{tabular}
\end{table}

\subsection{Results and Discussion}

The reliability of the proposed model largely hinges on accurate detection of the release frame and the precision of the pose estimation component. To assess this, we experimented with two different pose estimation libraries and various models for release frame detection. The results, summarized in Table \ref{tab:results}, were evaluated using seven performance metrics.

Initial experiments with Convolutional Neural Networks (CNNs) for release frame detection did not yield satisfactory results, resulting in low overall accuracy when incorporated into the pipeline. However, significant performance improvements were observed with object detection models. Only object detection models yielded a positive Matthews Correlation Coefficient (MCC), indicating that predictions were neither random nor inversely correlated with actual outcomes.

The experimental results demonstrate that conventional CNN-based release frame detectors, namely VGG-16, VGG-19, and ResNet-50, are not suitable for this task. Regardless of the pose estimation framework, these models achieved low classification accuracy and produced negative MCC values, indicating poor correlation between predicted and actual classes. These findings suggest that image classification models are inadequate for accurately localizing the precise ball release moment, which subsequently degrades the overall bowling action classification performance.

Replacing CNN-based methods with object detection models resulted in substantial performance improvements. When combined with OpenPose, YOLOv8 achieved an accuracy of 72.73\%, while RT-DETR further improved the accuracy to 75.00\%, increasing the Recall from 71.43\% to 85.71\% and the MCC from 0.4485 to 0.4781. Although OpenPose provided reasonable pose estimation performance, its relatively slower inference speed makes it less suitable for real-time deployment.

MediaPipe consistently outperformed OpenPose across all release frame detectors due to its lightweight architecture and efficient landmark tracking. When paired with YOLOv8, MediaPipe achieved an accuracy of 83.33\% with an F1-score of 87.50\%. The best overall performance was obtained by combining MediaPipe with RT-DETR, achieving an accuracy of 91.96\%, precision of 87.50\%, recall of 100\%, F1-score of 93.33\%, TPR of 100\%, FPR of 20\%, and an MCC of 0.8367. These results indicate that the proposed combination provides the most reliable discrimination between legal and illegal bowling actions.

The perfect recall obtained by the MediaPipe--RT-DETR pipeline indicates that all illegal bowling actions in the test set were correctly identified. From an umpiring perspective, this characteristic is highly desirable because failing to detect an illegal bowling action may directly influence the fairness of a match. Although the model produces a small number of false positives, reflected by a precision of 87.50\%, these cases can subsequently be verified through biomechanical assessment, whereas undetected illegal deliveries may have irreversible consequences during gameplay.
\begin{figure} [h]
    \centering
    \includegraphics[width=\linewidth]{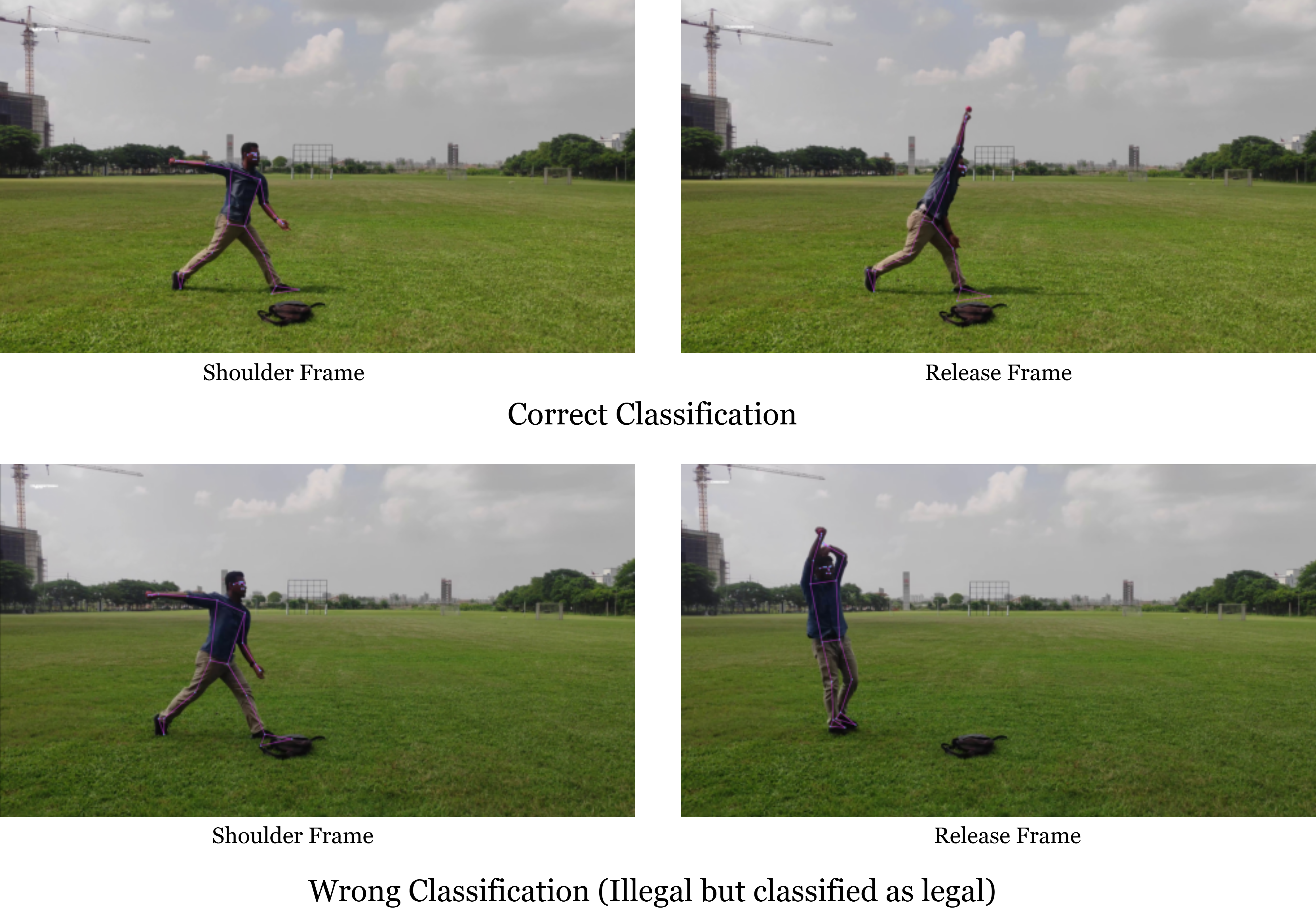}
    \caption{Visualization of correct and incorrect classification}
    \label{fig:visual}
\end{figure}

Figure~\ref{fig:visual} illustrates examples of both correctly and incorrectly classified deliveries. In the correctly classified example, both the shoulder frame and release frame were accurately detected, allowing precise elbow angle estimation and correct classification. Conversely, the misclassified example demonstrates that an inaccurately detected release frame propagates errors to the elbow angle calculation, ultimately leading to an incorrect prediction. These examples emphasize that reliable release frame localization is fundamental to the success of the proposed pipeline.

While high recall ensures that all illegal actions are detected, the challenge lies in minimizing false positives without increasing the risk of false negatives, which can have significant consequences for players and match outcomes. In this case, a false negative occurs when a legal delivery is incorrectly classified as illegal. If the umpire deems the delivery illegal, the bowler could be unfairly penalized with a no-ball or have a valid wicket revoked. Frequent false negatives could lead to disputes, affecting match strategy, momentum, and outcomes. Teams may lose runs, or a crucial wicket may be wrongfully disallowed. Therefore, minimizing false negatives should be a priority in the upcoming research.

While the results indicate promising performance, it is important to recognize limitations. The accuracy of the proposed model is highly dependent on both the quality of frame extraction and pose estimation accuracy. Further testing across a more diverse dataset and different environments is necessary to validate its robustness for various bowling styles and match conditions. Future work will focus on fine-tuning the model and improving frame detection and pose estimation components to enhance reliability for real-time applications.

\begin{table*}[h]
    \centering
    \caption{Performance comparison of different models}
    \begin{scriptsize}
    \begin{tabular}{|p{1.5cm}|p{1.5cm}|l|l|l|l|l|l|l|}
    \hline
        Pose estimation for elbow angle& Release frame detection & \multicolumn{7}{c|}{Results} \\ \cmidrule{3-9}
         & & Accuracy  & Precision  & Recall  & F1-Score & TPR  & FPR  & MCC \\ \hline
        {OpenPose} & VGG-16 & 0.4167 & 0.5 & 0.4286 & 0.4615 & 0.4286 & 0.6 & -0.169 \\
        ~ & VGG-19 & 0.3333 & 0.4444 & 0.5714 & 0.5 & 0.5714 & 1 & -0.488 \\
        ~ & ResNet-50 & 0.5 & 0.5556 & 0.7143 & 0.625 & 0.7143 & 0.8 & -0.0976 \\
        ~ & YOLOv8 & 0.7273 & 0.8333 & 0.7143 & 0.7692 & 0.7143 & 0.25 & 0.4485 \\
        ~ & RT-DETR& 0.75 & 0.75 & 0.8571 & 0.8 & 0.8571 & 0.4 & 0.4781 \\ \hline
        {\textbf{Mediapipe}} & VGG-16 & 0.4167 & 0.5 & 0.4286 & 0.4615 & 0.4286 & 0.6 & -0.169 \\ 
        ~ & VGG-19 & 0.3333 & 0.4444 & 0.5714 & 0.5 & 0.5714 & 1 & -0.488 \\ 
        ~ & ResNet-50 & 0.5 & 0.5556 & 0.7143 & 0.625 & 0.7143 & 0.8 & -0.0976 \\ 
        ~ & YOLOv8 & 0.8333 & 0.875 & 0.875 & 0.875 & 0.875 & 0.25 & 0.625 \\ 
        ~ & \textbf{RT-DETR} & \textbf{0.9196} & \textbf{0.875} & \textbf{1} & \textbf{0.9333} & \textbf{1} & \textbf{0.2} & \textbf{0.8367} \\ \hline
    \end{tabular}
    \end{scriptsize}
    \label{tab:results}
\end{table*}

\section{Conclusion}
\label{sec:conclu}
In this work, we introduced an approach for detecting illegal bowling actions using computer vision. Our system extracts two key frames—the shoulder frame and the release frame—to measure the change in the bowling arm's angle. If this difference exceeds a predefined threshold (\SI{15}{\degree}), the action is flagged as illegal. The proposed system was evaluated on a dataset labeled by human experts, achieving a true positive rate of 100\% and an accuracy of 90\% in distinguishing between legal and illegal actions. These results suggest that the system shows promise as a potential tool for real-time bowling action detection.

Despite these promising findings, the current work has limitations that should be addressed in future studies. The dataset lacks diversity, as it does not include female bowlers, bowlers with side-arm actions, or left-handed bowlers, potentially limiting the generalizability of the system. Additionally, since no other computer vision-based methods specifically address illegal bowling detection, a direct comparison to evaluate relative effectiveness is currently unavailable. Furthermore, 20\% of illegal actions were incorrectly classified as legal, indicating that the system could benefit from further refinement to enhance its reliability.

Since the proposed method is lightweight, it can be deployed both locally and in the cloud. Additionally, as the model takes only a few seconds for inference, it can be integrated into live matches seamlessly. It system has potential real-world applications in live broadcast analytics, automated umpiring assistance, and post-match analysis for performance evaluation and training. The future works will focus on expanding the dataset to include a wider variety of bowlers, which may help improve the system’s adaptability across different bowling styles and conditions. Enhancements to the model will also be explored to reduce the rate of false negatives, ultimately increasing the accuracy and dependability of the system for live match scenarios.
\section*{Declarations}
\textbf{Conflict of Interests:} On behalf of all authors, the corresponding author states that there is no conflict of interest.\\
\textbf{Funding:} No external funding is available for this research.\\
\textbf{Data and Code Availability}
We have shared the dataset and source code for the solution in the GitHub repository \cite{github_repo}.
\bibliographystyle{unsrt}
\bibliography{cas-refs}

\end{document}